\documentclass{article}
\usepackage{graphicx}
\usepackage{float}
\usepackage{cite}
\usepackage{afterpage}
\usepackage[margin=2.5cm]{geometry}
\usepackage{array}
\usepackage{booktabs}
\usepackage{threeparttable}%
\usepackage{multirow}
\usepackage[colorlinks=true, linkcolor=blue, citecolor=blue, urlcolor=blue]{hyperref}

\title{Multimodal MRI-Based Alzheimer’s Disease Diagnosis with Transformer-based Image Synthesis and Transfer Learning}
\author{Jason Qiu \\
Marvin Ridge High School \\}
\date{}

\begin{document}

\maketitle

\begin{abstract}
\noindent
Alzheimer’s disease (AD) is a progressive neurodegenerative disorder in which pathological changes 
begin many years before the onset of clinical symptoms, making early detection essential for timely 
intervention. T1-weighted (T1w) MRI is routinely used in clinical practice to identify macroscopic 
brain alterations such as cortical thinning and hippocampal atrophy, but these changes typically 
emerge relatively late in the disease course. Diffusion MRI (dMRI), in contrast, is sensitive to 
earlier microstructural abnormalities by probing water diffusion in brain tissue. Diffusion tensor 
imaging (DTI)–derived metrics, including fractional anisotropy (FA) and mean diffusivity (MD), 
provide complementary information about white matter integrity and neurodegeneration. However, 
dMRI acquisitions are time-consuming and susceptible to motion artifacts, limiting their routine 
use in clinical populations.

In this work, I propose a transformer-based image synthesis framework that predicts FA and MD maps 
directly from T1-weighted MRI, enabling access to microstructural diffusion information without 
additional scan time. A 3D TransUNet architecture is employed to capture long-range spatial 
dependencies and learn the complex relationship between macrostructural anatomy and diffusion-derived 
microstructure. The synthesized FA and MD maps are subsequently combined with T1w MRI as multi-modality 
input for Alzheimer’s disease classification. Quantitative evaluation shows that the proposed model 
accurately reconstructs diffusion parameters from T1w MRI, achieving low voxel-wise errors, high 
structural similarity, and strong correlation with reference FA and MD maps. MD prediction demonstrates 
particularly high fidelity, indicating robust preservation of biologically meaningful microstructural 
features. When incorporated into a multi-modality diagnostic model, the synthesized diffusion features 
significantly improve classification accuracy compared to a T1w-only baseline. The performance gain 
is especially pronounced for mild cognitive impairment (MCI), a clinically important and challenging 
stage for early diagnosis.

Overall, this study demonstrates that high-quality diffusion microstructural information can be 
inferred from routinely acquired T1-weighted MRI, effectively transferring the benefits of 
multi-modality imaging to settings where diffusion data are unavailable. By reducing scan time 
while preserving complementary structural and microstructural information, the proposed approach 
has the potential to improve the accessibility, efficiency, and accuracy of Alzheimer’s disease 
diagnosis in clinical practice. \\

\textbf{Originality Statement}
This work is original and authored solely by Jason Qiu. All analyses, models, figures, and interpretations presented herein are the author’s own work. The research is based on publicly available datasets (HCP, ADNI), and all methods and results are independently developed by the author.

\end{abstract}

\section{Introduction}
Alzheimer’s Disease (AD) is a major concern in our ageing society. It results from neurodegeneration and leads to deficits in memory, cognitive function, and behavior. AD also increases the risk of other diseases and disorders such as type 2 diabetes and glaucoma through comorbidity. Globally, there are over 55 million people with AD or AD related diseases \cite{29} and this number is expected to increase as the number of older populations increase. There is currently no cure for AD. Therefore, because neurodegeneration starts many years before cognitive symptoms, early detection of the disease is crucial for timely intervention before irreversible damage happens.

A promising method that has the potential to achieve early AD detection is magnetic resonance imaging (MRI). MRI is a noninvasive tool for detecting brain structural changes typically seen in AD. For example, T1-weighted (T1w) MRI can provide excellent contrast between gray and white matter, and is routinely used in clinical practice to detect AD signs such as cortical thinning and hippocampus atrophy \cite{3,7,30}. However, macrostructural changes detectable with T1w MRI also happen years after the start of neurodegeneration and neuronal loss. As a result, irreversible neuronal damage might have already happened by the time T1w MRI detects changes. 

Diffusion MRI (dMRI) has the potential to offer higher sensitivity to earlier disease changes. dMRI reveals even finer level, cellular information by sensitizing the signal to water diffusion. and can provide information about cellular morphology and density \cite{1,2}. The cellular-level microstructural changes are complementary to the macrostructural changes revealed by T1w \cite{31}. Two commonly derived metrics from diffusion tensor imaging (DTI) are fractional anisotropy (FA) and mean diffusivity (MD). FA reflects the directional coherence of water diffusion along white matter fibers. A decrease in FA often indicates demyelination or axonal degeneration \cite{18}. MD measures the overall magnitude of water diffusion. MD increases often reflects tissue loss, edema, or neurodegeneration \cite{20}. These metrics are widely used in neuroscience and clinical studies including studies of stroke, neurodegenerative diseases, and white matter abnormalities. 

The complementary macrostructural (T1w) and microstructural (dMRI) information motivates the use of multi-modality input for earlier and more accurate AD diagnosis. Previous studies have shown that multi-modality inputs including T1w and dMRI effectively improve the accuracy of AD diagnosis models \cite{32,33}. However, the long scan time required to collect multi-modality data pose challenges for their clinical use. T1w MRI is routinely acquired in clinical scans, taking about 8 minutes. Both FA and MD are computed by fitting a diffusion tensor model to a series of diffusion-weighted images (DWIs), which requires acquiring multiple images with different diffusion directions and b-values. This makes conventional DTI scans time-consuming (typically more than 9 minutes) and prone to motion artifacts. This is especially problematic in AD patients who find it even more difficult to tolerate long scans compared to healthy people. The recent success of deep learning has shown promise in addressing these challenges. Previous work has leveraged deep learning to synthesize dMRI from T1w MRI, mostly through the use of Convolutional Neural Networks (CNN) \cite{6,34,35}. This represents a promising direction as the accurate synthesis of dMRI could eliminate the required scan time for multi-modality input while still harnessing the accuracy benefits. They showed promising results and suggested that structural MRI contains sufficient information to approximate diffusion properties. However, the inherent limitation of CNNs in capturing the long-range dependencies of images limits the performance of these methods. Furthermore, their generalizability to clinical datasets remains uncertain.

In this study, I propose to leverage transformers through the use of a 3D TransUNet to capture the long-range dependencies in the images, and to achieve high-fidelity FA and MD prediction from T1w input. The synthesized FA and MD are jointly used with T1w input as multi-modality input to improve AD diagnosis accuracy. In addition to the diagnosis of advanced AD patients, I also evaluated this strategy’s efficacy in identifying patients showing early cognitive deficits, known as mild cognitive impairment (MCI). The proposed image synthesis model produces high-quality FA and MD and shows generalizability to clinical datasets with AD patients. Further analysis shows that predicted FA and MD maps can improve the diagnosis accuracy of AD through multimodality inputs compared to single-modality input with only T1w. These results highlight the promise of deep learning-assisted image synthesis in advancing AD diagnosis accuracy, potentially enabling earlier detection and better patient care for AD. 

\section{Related Work}

\subsection{Deep Learning}

3D CNNs provide strong local modeling but lack global context. Transformer-based models can capture long-range relationships but require high memory \cite{28}. U-Net \cite{9} is one of the most widely used architectures in medical image analysis. Its encoder–decoder structure with skip connections enables effective fusion of semantic features and spatial details, which makes it a strong baseline in segmentation and image-to-image prediction tasks \cite{8,12,22}.

TransUNet \cite{10} extends U-Net by incorporating a Transformer module into the encoder. U-Net focuses on local feature extraction through convolutions, while TransUNet adds global self-attention to capture long-range dependencies and richer anatomical patterns. This hybrid CNN–Transformer design has demonstrated improved performance over CNN-only architectures in various medical imaging applications \cite{10}.

\subsection{Loss Functions}

L1 is widely adopted in quantitative MRI regression tasks because it provides stable optimization and preserves boundary information better than L2 \cite{14,15}. SSIM can improve the structure, but consumes a lot of 3D memory \cite{5}.

\subsection{Patch-based learning}

Due to GPU constraints, 3D networks often rely on patch-based training \cite{21,16}. MRI scans often show considerable variation in intensity across scanners and acquisition settings, so a stable normalization step—commonly z-score normalization—is usually needed to keep the data comparable \cite{17}.

\section{Methods}

A TransUNet architecture was used to predict voxel-wise FA and MD maps from T1-weighted MRI. The network consists of convolutional layers for local feature extraction with Transformer layers to capture long-range dependencies. Each convolutional block has two consecutive convolutional layers. The base channel number was set to 32, and input volumes were divided into 3D patches of size 646464.

A Convolutional Neural Network architecture was used to diagnosis Alzheimer’s Disease from original T1-weighted MRI images and synthesized FA and MD maps. The selection of a CNN was aimed to automatically learn hierarchical features, reducing parameters, providing an accurate model for classification. The base channel number was set to 30 and the output 

\subsection{Hardware}

All experiments were carried out on a workstation equipped with an Intel Core i5-8600 CPU (up to 4.3 GHz), an NVIDIA GeForce RTX 3070 GPU with 8 GB of VRAM, and 32 GB of DDR4 RAM. The operating system was Windows 11 Home, and all models were implemented in PyTorch 3.12.10 with CUDA 13.0 support.

\subsection{Synthesis Model}

\subsubsection{Dataset}

For pre-training, I utilized data from the Human Connectome Project (HCP) \cite{4}. The raw HCP data were first downloaded and subsequently processed to extract FA, MD feature maps and T1 weighted contrast images. This preprocessing aligned T1w images, corresponding FA and MD maps, and binary brain masks for each subject. These processed images were used as inputs and targets for the subsequent 3D neural network training. The acquisition time of T1w images is roughly 7 minutes, 40 seconds \cite{36}.

\subsubsection{Model}

I employed a 3D TransUNet architecture, which consists of a baseline UNet model with a Transformer-based bottleneck. The baseline UNet model consists of a convolutional encoder, and a convolutional decoder. The convolutional layers are implemented as Double Convolution blocks, each followed by batch normalization and ReLU activation. The network uses a base channel size of 32.

\begin{figure}[H]
    \centering
    \includegraphics[width=1\linewidth]{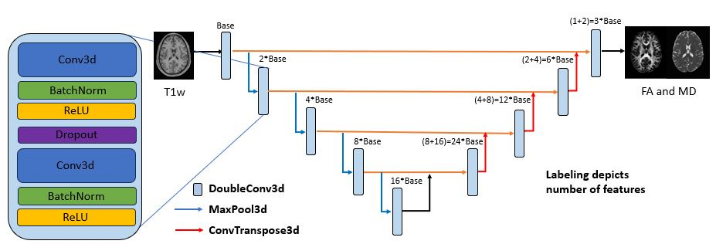}
    \caption{Architecture for UNet Model}
    \label{Architecture for UNet Model}
\end{figure}

For multi-task learning, the decoder is designed with two separate output heads to predict FA and MD maps simultaneously. The network takes a single-channel T1-weighted (T1w) MRI patch of size (1, 64, 64, 64) as input and outputs a two-channel FA and MD map of size (2, 64, 64, 64).

\begin{figure}[H]
    \centering
    \includegraphics[width=1\linewidth]{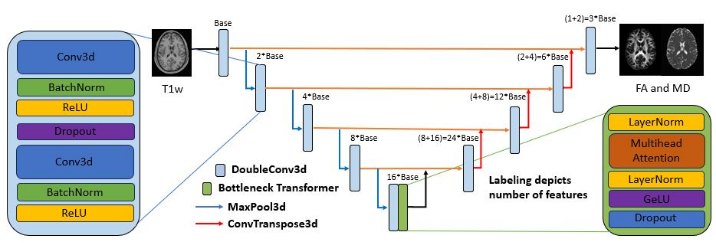}
    \caption{Architecture for TransUNet Model}
    \label{Architecture for TransUNet Model}
\end{figure}
The transformer model that was used for synthesis consisted of an additional Transformer layer at the bottleneck of the UNet. The bottleneck consists of normalization followed by a multi-head attention layer, another normalization, ReLU, and a dropout of 0.5. 

\subsubsection{Training Details}

The dataset, consisting of 100 subjects, was split into 60 training, 20 validation, and 20 testing samples. The network was trained using the Adam optimizer with a learning rate of 1e-4 and L1 as loss function. Each training batch contained 10 patches of size 64\textsuperscript{\textsuperscript{3}}. Training was run for 100 to 150 epochs, with early stopping using a patience of 10 epochs. Automatic mixed precision (AMP) was not used during the training.

\subsubsection{Evaluation Metrics}

The performance of the synthesis model was assessed using several commonly used quantitative metrics:

\begin{itemize}
    \item \textbf{Mean Squared Error (MSE) }– measures the average squared difference between the pre-dicted and reference values.
    \item \textbf{Mean Absolute Error (MAE) }– quantifies the average absolute prediction error.
    \item \textbf{Root Mean Squared Error (RMSE) }– reflects the square root of MSE and penalizes larger errors more strongly.
    \item \textbf{Structural Similarity Index (SSIM) }– evaluates similarity in structural patterns and image contrast.
    \item \textbf{Pearson Correlation Coefficient (r) }– measures the linear correspondence between predicted and reference maps.
\end{itemize}

\subsubsection{Transfer Learning}

The model trained on HCP was applied to ADNI where native FA and MD might be unavailable.

\subsection{Prediction Model}

\subsubsection{Dataset}

For the second phase prediction model, data was obtained from Alzheimer’s Disease Neuroimaging Initiative (ADNI). To ensure consistency all data obtained were preprocessed, 3D, and had a field strength of 3 tesla. A total of 800 subjects, 400 AD and 400 Normal Control (NC), were downloaded and subsequently processed to extract FA and MD maps, T1-weighted contrast images, and binary brain masks for each subject The processed images were first trained using the synthesis model and subsequently used as input for the prediction model. The acquisition time of all dMRI data is roughly 9 minutes and 50 seconds, substantially longer than acquiring T1w images only. 

For the input of the prediction model, each 280x280x180 image was sliced into 2D 280x280 slices to and 30 centrally located slices were taken in an every-other-slice format from each T1w, FA, and MD image \cite{16}. Theoretically a 3D model would have provide better performance by capturing more complex spatial relationships. However, due to the hardware limitation that the GPU does not have enough RAM to run a 3D model, I selected a 2D model instead. Although a 2D model was used, all training, validation, and testing data was divided subject-wise instead of slice-wise to more closely mimic a 3D model despite limitations. 

\subsubsection{Model}

A standard 2D Convolutional Neural Network architecture consisting of convolutional layers and fully connected layers was implemented. The input consisted of 3 features (FA, MD, and T1w) followed by four convolutional layers consisting of down sampling, batch normalization, ReLU, and pooling. The fully connected layers consist of an Adaptive Average Pool to reduce the number of features for the linear layer and three subsequent layers with dropout of 0.5 and ReLU. The final output of the model consists of 1 feature passed through a sigmoid function to classify AD or NC. See Appendix B for model diagrams. 

Two diagnosis strategies have been evaluated: NC vs. AD classification and NC vs. MCI vs. AD classification. To further evaluate the clinical benefits of the synthesized FA and MD modalities, I implemented a three-classification diagnosis model that categorizes patients as AD, NC, and Mild Cognitive Impairment (MCI). MCI, defined as the in-between stage as normal cognitive functions and Alzheimer’s Disease, is the most difficult dementia-related condition to diagnose due to and a lack of properly developed cognitive tests. Current cutting-edge models can only perform binary classification of MCI and NC at 73\% showing the difficulty of MCI prediction not only by medical professionals but also by machine learning. Oftentimes, MCI patients may not find the incentive to seek treatment due to symptoms perceived as insignificant. This has resulted in an estimated of only 8\% of MCI patients to be properly diagnosed \cite{37}, which poses a significant risk to public health as MCI can progress into AD. 

\subsubsection{Training Details}

The dataset for the prediction model consisted of 400 subjects split into 320 training, 40 validations, and 40 testing for both AD and NC subjects. The network was trained using the Adam optimizer with a learning rate of 5e-4 and BCEWithLogitsLoss as the loss function. Each training batch consisted of 30 slices. Training generally ran for 15-20 epochs with early stoppage using a patience of 10 epochs. 

Augmentation consisting of random flip and intensity jitter was used on the training data only. Intensity jitter with a scale of 0.02 was used for all AD patients’ data and intensity jitter with a scale of 0.0775 was used for all NC patients’ data

\subsubsection{Evaluation Metrics}

The performance of the prediction model was assessed using several commonly used quantitative metrics:

\begin{itemize}
    \item \textbf{Accuracy} – Measures the proportion of all predictions that are correct
    \item \textbf{Precision} - The proportion of predicted positive cases that are actually positive
    \item \textbf{Recall} – The proportion of actual positive cases that are correctly identified  
    \item \textbf{F1 Score} – measures the harmonic mean of precision and recall and effectively reflects the balance between the two values. 
    \item \textbf{Subject-Level AUC} – measures the model's ability to distinguish between classes with 1 indicating perfect classification and 0.5 indicating random guess. 
\end{itemize}

To determine whether each image was categorized correctly as AD or NC, the cumulative prediction of all of the slices of each patient was evaluated. Having more than 20\% of slices predict AD resulted in a patient being categorized as AD and having less than 20\% categorized the patient as NC. 

\section{Experiments and Results}

The proposed synthesis model achieves high accuracy in predicting diffusion parameters (FA and MD) from T1-weighted MRI. Training is performed using patches of size 64\textsuperscript{\textsuperscript{3}}, and an early stopping strategy is employed to mitigate overfitting.

\subsection{Synthesis Model}

\subsubsection{Prediction Accuracy}

\begin{table}[H]
\centering
\caption{Evaluation Metrics for FA and MD Predictions.}
\begin{tabular}{l l l}
\hline
\textbf{\textbf{Metric}} & \textbf{\textbf{FA Prediction}} & \textbf{\textbf{MD Prediction}} \\
\hline
MSE & 0.00221 & 0.00127 \\
MAE & 0.01255 & 0.00869 \\
RMSE & 0.04695 & 0.03552 \\
SSIM & 0.93541 & 0.96678 \\
Pearson \textit{\textit{r}} & 0.94089 & 0.97224 \\
\hline

\end{tabular}

\end{table}

The model achieves high accuracy in predicting both FA and MD which demonstrates its ability to effectively reconstruct white matter microstructural information. For FA, voxel-level reconstruction errors on the test set were low (MSE = 0.00221, MAE = 0.01255, RMSE = 0.04695), and the predicted maps exhibited high structural similarity with the ground truth (SSIM = 0.93541) \cite{5}, which indicated the model preserves white matter architecture and local tissue contrast.

For MD, model performance was even higher (MSE = 0.00127, MAE = 0.00869, and RMSE = 0.03552). The predicted MD maps showed a very strong correlation with the ground truth (Pearson r = 0.97224) and high structural fidelity (SSIM = 0.96678). This reflected excellent preservation of both intensity and spatial structure.

\subsubsection{Training Stability}

The training and validation loss curves show that the model converges smoothly, and the Early Stop strategy effectively avoids overfitting. The predictions of FA and MD stabilize after a relatively small number of training epochs which indicates that TransUNet can learn meaningful features even on small sample data.

\begin{figure}[H]
    \centering
    \includegraphics[width=0.75\linewidth]{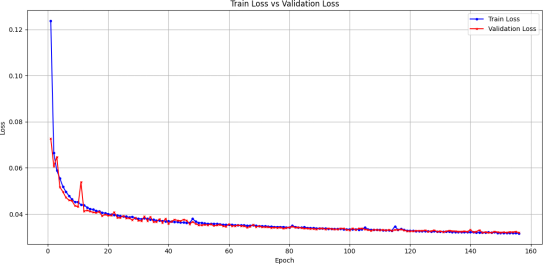}
    \caption{The training and validation loss curves.}
    \label{The training and validation loss curves}
\end{figure}

\subsubsection{Visual Evaluation}

When comparing the predicted FA and MD maps with the ground-truth DTI images, I noticed that the major white-matter pathways were still clearly recognizable in the predictions. Some of the finer details appeared a bit smoother, but the overall spatial patterns remained very close to the true images. This visual similarity aligns well with the quantitative evaluation.

\begin{figure}[H]
    \centering
    \includegraphics[width=0.6\linewidth]{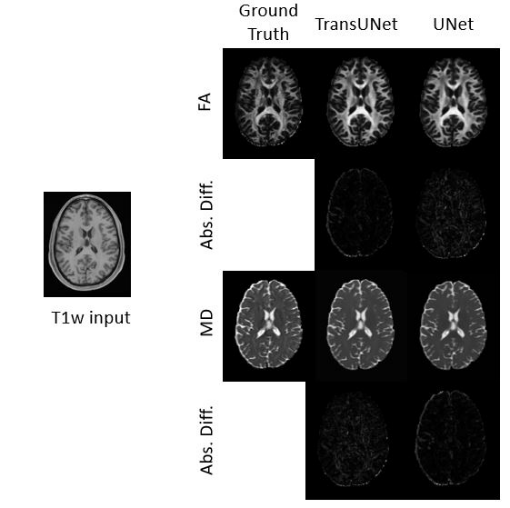}
    \caption{Prediction Results of FA and MD for TransUNet and UNet}
    \label{Prediction Results of FA and MD for TransUNet and UNet}
\end{figure}

Visually, the TransUNet synthesis is superior for both FA and MD compared to a UNet synthesis as both categories of parametric maps show less absolute difference leading to closer resemblance to the ground truth. The FA prediction by the TransUNet shows more precise boundaries between gray and white matter compared to the UNet and the MD prediction by the TransUNet shows more detailed information towards the borders of the parametric map. 

\subsubsection{Ablation Studies}

To assess the contributions of different components in the model, a series of ablation experiments were performed \cite{9,13}. The factors examined included single- vs. double-convolution blocks, removing the Transformer block, different normalization methods and L1 vs L2 loss. Results indicated that double-convolution blocks improved feature extraction \cite{9}, while the chosen patch size and channel configuration achieved a balance between prediction accuracy and GPU memory consumption. Including the Transformer improved the model’s ability to capture long-range dependencies \cite{13}.

\paragraph{Transformer Module, Upsampling Depth, and Convolutional Block}

To assess the contribution of each architectural component, several model variants were evaluated, including the full TransUNet, a UNet-only version with the Transformer module removed, and a reduced-depth model in which the encoder and decoder were shortened from four to two levels. I also compared single- versus double-convolution blocks. Incorporating the Transformer module improved the model’s ability to capture long-range dependencies, which is important for representing global anatomical structure \cite{10,26}. Increasing the encoder/decoder depth provided richer contextual information but required more GPU memory \cite{9}. Double-convolution blocks enabled more expressive local feature extraction and better preservation of structural boundaries, leading to improved voxel-level prediction performance \cite{24}.

\paragraph{Input Normalization}

I examined z-score normalization (main approach) versus min-max normalization and no normalization. Z-score normalization helps reduce intensity variability across scanners and subjects, which stabilizes the training process and improves the model’s ability to generalize \cite{24}.

\paragraph{Loss Function}

I compared L1 loss (main method), L2 loss, and L1+SSIM. L1 provided stable optimization and better edge preservation. SSIM improved structure but increased 3D memory usage.

\paragraph{Summary}

Overall, the ablation study confirms that multi-task learning, Transformer modules, z-score normalization, and a L1 loss function are critical to high-quality FA and MD synthesis \cite{26,24}. Further ablation studies were conducted on Multi task vs single task prediction, input normalization, model width, and patch size. (See Appendix Section 1)


\begin{table}[H]
\centering
\caption{Ablation Metrics for FA and MD Predictions.}
\begin{tabular}{l l l l l l l}
\hline
\multicolumn{2}{l}{\textbf{\textbf{Ablation}}} & \textbf{\textbf{MAE}} & \textbf{\textbf{MSE}} & \textbf{\textbf{RMSE}} & \textbf{\textbf{SSIM}} & \textbf{\textbf{Pearson}} \\
\hline
\multirow{2}{*}{\textbf{\textbf{Proposed model}}} & FA & 0.05380 & 0.01005 & 0.09797 & 0.73281 & 0.87576 \\
 & MD & 0.03700 & 0.00600 & 0.07597 & 0.79816 & 0.91832 \\
\multirow{2}{*}{Transformer removed} & FA & 0.07118 & 0.01676 & 0.12401 & 0.80404 & 0.62965 \\
 & MD & 0.05156 & 0.01172 & 0.10062 & 0.85727 & 0.70767 \\
\multirow{2}{*}{No normalization} & FA & 0.07318 & 0.02098 & 0.12850 & 0.80355 & 0.57761 \\
 & MD & 0.05491 & 0.01431 & 0.11318 & 0.82402 & 0.63036 \\
\multirow{2}{*}{L2 loss} & FA & 0.08462 & 0.01630 & 0.12615 & 0.78627 & 0.45378 \\
 & MD & 0.07066 & 0.01454 & 0.11729 & 0.80471 & 0.47461 \\
\hline
\end{tabular}

\end{table}

\subsubsection{Image Synthesis on ADNI}

The model trained on HCP data synthesized high-quality images on ADNI data. HCP consists of well-organized pre-processed data, while ADNI consists of largely clinical data obtained through various methods \cite{11,23}. The ability for the synthesis model to perform exceptionally well on ADNI when trained on HCP data demonstrates the generalizability in which the model can learn different features. 

\begin{figure}[H]
    \centering
    \includegraphics[width=0.6\linewidth]{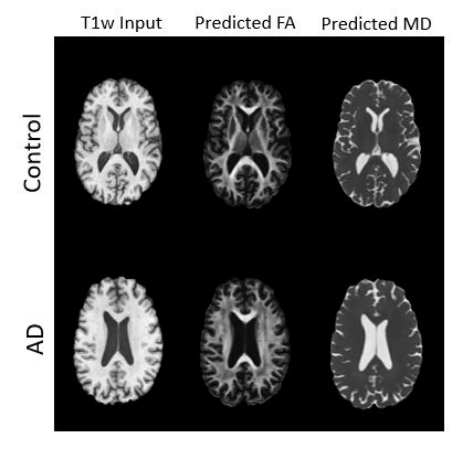}
    \caption{Input (T1w) Image vs Output (FA + MD) Images for AD and NC patients}
    \label{Input (T1w) Image vs Output (FA + MD) Images for AD and NC patients}
\end{figure}

\subsection{Prediction Models}

\subsubsection{Prediction Accuracy}
\begin{table}[H]
\centering
\caption{Evaluation Metrics for FA and MD Predictions.}
\begin{tabular}{l l l}
\hline
\textbf{\textbf{Metric}} & \textbf{\textbf{NC}} & \textbf{\textbf{AD}} \\
\hline
Accuracy & 0.850 & 0.825 \\
Precision & 0.829 & 0.846 \\
Recall & 0.850 & 0.825 \\
F1 Score & 0.840 & 0.835 \\
\hline

\end{tabular}

\end{table}

\begin{table}[H]
\centering
\caption{Confusion Matrix of Predictions}
\begin{threeparttable}
\begin{tabular}{l l l}
\hline
 & \textbf{\textbf{Predicted as AD}} & \textbf{\textbf{Predicted as NC}} \\
\hline
AD & 34 & 6 \\
NC & 7 & 33 \\
\hline

\end{tabular}
    \begin{tablenotes}
    \footnotesize  
    \item[*] Overall Accuracy: 0.8375
    \item[*] Subject-Level AUC: 0.9069
    \end{tablenotes}
\end{threeparttable}
\end{table}

The model achieves high accuracy in predicting both NC and AD demonstrating the ability for the model to detect changes in neural composition and structure that are resulted from AD even if these changes may not be prevalent in all slices. The higher precision of 0.846 observed for the AD prediction suggests that the model makes relatively few false positive predictions, indicating conservative behavior when identifying disease cases. Conversely, the very high recall value of 0.9069 for predicting NC indicates that the model is able to confidently diagnose healthy patients. This reflects a trade-off between precision and recall where the model mildly prioritizes confidence in AD predictions over sensitivity. 

However, the similar F1 scores of 0.840 for NC and 0.835 and a very strong subject-level AUC of 0.9069 shows the model has an appropriate balance between the two classes despite the indicator of a slightly conservative model by the precision and recall scores independently.

\subsubsection{Training Stability}

The training curve shows a smooth increasing curve indicating that the prediction model is correctly and consistently learning features from T1w, FA and MD. The validation loss curve shows an increasing line of best fit with fluctuations decreasing as epochs increase indicating that the model has learned all the appropriate features it is capable of learning. The early stoppage strategy effectively avoids overfitting by stopping the training process when the training loss curves flattens out and the validation loss begins to converge.
\begin{figure}[H]
    \centering
    \includegraphics[width=0.75\linewidth]{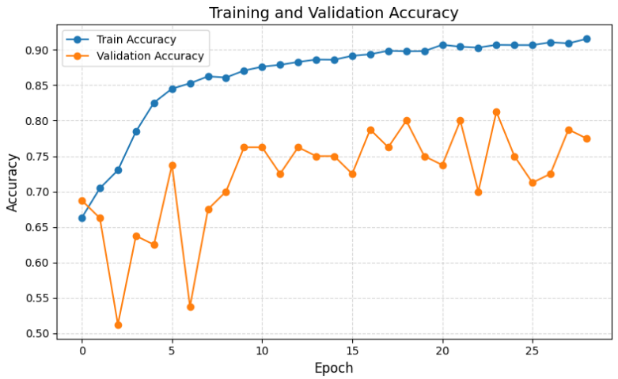}
    \caption{Training vs Validation Accuracy of Prediction Model Graph}
    \label{Training vs Validation Accuracy of Prediction Model Graph}
\end{figure}

\subsubsection{Ablation Studies}

To assess the effectiveness of the multi-modality prediction model, an ablation study was conducted against a single-modality prediction model using original T1w data as input. To ensure consistency for comparison, all other aspects of the single-modality model that were permitted to be the same as the multi-modality were kept constant. These included the format of the input as slices, convolutional layers, the fully connected layers, loss function, optimizer, and training, validation, and testing data. 

\begin{table}[H]
\centering
\caption{Evaluation Metrics for FA and MD Predictions of Multi-Modality Model.}
\begin{tabular}{l l l}
\hline
\textbf{\textbf{Metric}} & \textbf{\textbf{NC}} & \textbf{\textbf{AD}} \\
\hline
Accuracy & 0.850 & 0.825 \\
Precision & 0.829 & 0.846 \\
Recall & 0.850 & 0.825 \\
F1 Score & 0.840 & 0.835 \\
\hline

\end{tabular}

\end{table}

\begin{table}[H]
\centering
\caption{Evaluation Metrics for FA and MD Predictions of Single Modality Model}
\begin{tabular}{l l l}
\hline
\textbf{\textbf{Metric}} & \textbf{\textbf{NC}} & \textbf{\textbf{AD}} \\
\hline
Accuracy & 0.750 & 0.825 \\
Precision & 0.811 & 0.767 \\
Recall & 0.750 & 0.825 \\
F1 Score & 0.779 & 0.795 \\
\hline

\end{tabular}

\end{table}

\begin{table}[H]
\centering
\caption{Confusion Matrix of Predictions for Multi-Modality Model}
\begin{tabular}{l l l}
\hline
 & \textbf{\textbf{Predicted as AD}} & \textbf{\textbf{Predicted as NC}} \\
\hline
AD & 34 & 6 \\
NC & 7 & 33 \\
\hline

\end{tabular}

\end{table}

\begin{table}[H]
\centering
\caption{Confusion Matrix of Predictions for Single Modality Model}
\begin{threeparttable}
\begin{tabular}{l l l}
\hline
 & \textbf{\textbf{Predicted as AD}} & \textbf{\textbf{Predicted as NC}} \\
\hline
AD & 30 & 10 \\
NC & 7 & 33 \\
\hline

\end{tabular}
    \begin{tablenotes}
    \footnotesize 
    \item[*] Subject-Level AUC Score for Multi-Modality: 0.9069
    \item[*] Subject-Level AUC Score for Single Modality Model: 0.8519
    \item[*] Overall Accuracy for Multi-Modality: 0.8375
    \item[*] Overall Accuracy for Single Modality Model: 0.7875
    \end{tablenotes}
\end{threeparttable}
\end{table}

The final results of the ablation study showed that the multi-modality model could diagnose AD at a 5.00\% higher accuracy than the single-modality model.

\subsection{Further Work through Three-Classification Prediction Model}

To further evaluate the clinical benefits of the synthesized FA and MD modalities, I implemented a three-classification diagnosis model that categorizes patients as AD, NC, and Mild Cognitive Impairment (MCI). MCI, defined as the in-between stage as normal cognitive functions and Alzheimer’s Disease, is the most difficult dementia-related condition to diagnose due to and a lack of properly developed cognitive tests. Current cutting-edge models can only perform binary classification of MCI and NC at 73\% showing the difficulty of MCI prediction not only by medical professionals but also by machine learning.

Oftentimes, MCI patients may not find the incentive to seek treatment due to symptoms perceived as insignificant. This has resulted in an estimated of only 8\% of MCI patients to be properly diagnosed \cite{37}, which poses a significant risk to public health as MCI can progress into AD.

In my further study, I built both a single modality three classification model with T1w only as input and a multi-modality three classification model with original T1w images and synthesized FA and MD maps. The multi-modality model was compared with the single modality prediction model to assess the effectiveness of dMRI features in predicting AD.

To ensure consistency for comparison, all aspects of the three-classification single-modality model and multi-modality model where kept constant as the binary-classification single-modality model and multi-modality model when permitted. These included the format of the input as slices, convolutional layers, optimizer, and training, validation, and testing data. However, for both three-classification model the loss function used was CrossEntropyLoss and the fully connected layer produced 3 channels as output instead of 1.

\begin{table}[H]
\centering
\caption{MCI Classification Performance: Single- vs. Multi-Modality}
\begin{tabular}{l l l l}
\hline
\textbf{\textbf{Metric}} & \textbf{\textbf{Single-Modality}} & \textbf{\textbf{Multi-Modality}} & \textbf{\textbf{Improvement}} \\
\hline
Accuracy & 0.425 & 0.550 & \textbf{\textbf{29.41\% ↑}} \\
Precision & 0.607 & 0.667 & \textbf{\textbf{9.88\% ↑}} \\
Recall & 0.425 & 0.603 & \textbf{\textbf{41.88\% ↑}} \\
F1 Score & 0.500 & 0.603 & \textbf{\textbf{20.60\% ↑}} \\
\hline

\end{tabular}

\end{table}

\begin{table}[H]
\centering
\caption{Evaluation Metrics for FA and MD Predictions of Multi-Modality Model.}
\begin{tabular}{l l l l}
\hline
\textbf{\textbf{Metric}} & \textbf{\textbf{NC}} & \textbf{\textbf{MCI}} & \textbf{\textbf{AD}} \\
\hline
Accuracy & 0.750 & 0.550 & 0.625 \\
Precision & 0.698 & 0.667 & 0.568 \\
Recall & 0.750 & 0.603 & 0.595 \\
F1 Score & 0.723 & 0.603 & 0.595 \\
\hline

\end{tabular}

\end{table}

\begin{table}[H]
\centering
\caption{Evaluation Metrics for FA and MD Predictions of Single Modality Model}
\begin{tabular}{l l l l}
\hline
\textbf{\textbf{Metric}} & \textbf{\textbf{NC}} & \textbf{\textbf{MCI}} & \textbf{\textbf{AD}} \\
\hline
Accuracy & 0.575 & 0.425 & 0.675 \\
Precision & 0.451 & 0.607 & 0.659 \\
Recall & 0.575 & 0.425 & 0.675 \\
F1 Score & 0.505 & 0.500 & 0.667 \\
\hline

\end{tabular}

\end{table}

\begin{table}[H]
\centering
\caption{Confusion Matrix of Predictions for Multi-Modality Model}
\begin{tabular}{l l l l}
\hline
 & \textbf{\textbf{Predicted as AD}} & \textbf{\textbf{Predicted as MCI}} & \textbf{\textbf{Predicted as NC}} \\
\hline
AD & 30 & 2 & 8 \\
MCI & 7 & 22 & 11 \\
NC & 6 & 9 & 25 \\
\hline

\end{tabular}

\end{table}

\begin{table}[H]
\centering
\caption{Confusion Matrix of Predictions for Single Modality Model}
\begin{threeparttable}
\begin{tabular}{l l l l}
\hline
 & \textbf{\textbf{Predicted as AD}} & \textbf{\textbf{Predicted as MCI}} & \textbf{\textbf{Predicted as NC}} \\
\hline
AD & 23 & 7 & 10 \\
MCI & 19 & 17 & 4 \\
NC & 9 & 4 & 27 \\
\hline

\end{tabular}
    \begin{tablenotes}
    \footnotesize
    \item[*] Overall Accuracy for Multi-Modality: 0.6417
    \item[*] Overall Accuracy for Single Modality Model: 0.5583
    \end{tablenotes}
\end{threeparttable}
\end{table}

The results of the three-classification model reveal promising results for the potential for MCI to be detected and singled out from AD and NC patients. An accuracy of 0.6417 for the multi-modality model shows the ability for mild neurodegeneration and changes to brain structure to be detected. Additionally, the 8.34\% increase in accuracy of the multi-modality model over the single-modality model shows the effectiveness of dMRI FA and MD in MCI diagnosis. Furthermore, for just MCI prediction, the accuracy of the multi-modality model predicts MCI at a 12.5\% higher accuracy than the single modality model- A value greater than the overall accuracy increase of the multi-modality model and far greater than the overall accuracy increase of 5.00\% for the binary multi-modality model. 

This drastic increase may be attributed to dMRI features’ abilities to detect smaller changes to the brain that precede major structural changes that result in memory loss \cite{38}. Specifically, FA and MD excel at detecting early disruptions in pathways connecting to the hippocampus, something that T1w along cannot, which may help explain the drastic increase in accuracy of the model. Additionally, the three-classification model further shows the applicability and accuracy of the synthesized dMRI features as they provide critical information for improved diagnosis. 

\section{Discussion}

In this study, I demonstrated DTI metrics FA and MD can be accurately synthesized from T1w MRI using a transformer-based model. The low voxel-wise reconstruction error and high structural similarity observed for FA indicate that the model effectively preserves white matter architecture and local tissue contrast. MD prediction achieved even stronger agreement with the reference maps, showing high structural fidelity and strong correlation with ground-truth MD values \cite{19}. The superior performance for both diffusion parameters highlights the benefit of the transformer architecture, whose ability to capture long-range spatial dependencies enables more accurate learning of the complex, non-local relationships between T1-weighted anatomy and diffusion-derived microstructural measures.

More broadly, this work represents an example of image quality transfer \cite{39}, in which the high-quality diffusion information available in large, well-controlled research datasets such as the HCP can be leveraged to enhance clinical datasets like ADNI. By learning a robust mapping between high-quality diffusion data and widely available T1-weighted MRI, the proposed approach transfers microstructural information that is otherwise inaccessible in many clinical settings. This strategy is particularly relevant for large-scale or retrospective studies where diffusion MRI is unavailable, incomplete, or degraded by motion artifacts.

The synthesized FA and MD maps were not only visually and quantitatively accurate, but also biologically meaningful, supporting their downstream clinical utility. This suggests that T1-weighted prediction of white matter microstructure may serve as a viable supplement to conventional DTI in scenarios involving limited scan time, patient intolerance, or missing diffusion acquisitions \cite{20}. Importantly, the proposed approach offers a non-invasive and low-cost alternative that retains clinically relevant microstructural information without requiring additional diffusion scanning.

The second phase of this study demonstrated that incorporating the predicted diffusion features into a multi-modality classification framework improved diagnostic performance for AD. While both the single-modality (T1w-only) and multi-modality models achieved relatively high accuracy (78.75\% and 83.75\%, respectively), the inclusion of FA and MD reduced the misclassification rate by 23.53\%. This improvement reflects the complementary nature of structural and diffusion MRI: T1-weighted MRI primarily captures macroscopic changes such as cortical atrophy and ventricular enlargement, whereas diffusion MRI is sensitive to subtle microstructural alterations in white matter that may precede overt structural loss.

These complementary strengths are particularly relevant for MCI, which remains one of the most challenging yet clinically critical stages to diagnose. Early microstructural changes are often detectable in diffusion metrics before pronounced gray matter atrophy becomes evident. Consequently, multi-modality approaches that incorporate diffusion-derived features are expected to provide greater benefit for MCI classification and early disease detection, supporting timely intervention when therapeutic strategies are most effective.

Overall, this study highlights the clinical significance of synthesizing diffusion MRI from T1-weighted images. By shortening scan times while retaining the benefits of multi-modality imaging, the proposed framework improves accessibility to advanced neuroimaging biomarkers without increasing patient burden. Leveraging machine learning to bridge structural and diffusion MRI enables more accurate, efficient, and affordable Alzheimer’s disease diagnosis, particularly in populations where extended or multi-sequence scanning is impractical.

\section{Conclusion}

This study shows that diffusion microstructural features (FA and MD) can be accurately synthesized from T1-weighted MRI using a transformer-based framework, enabling high-fidelity prediction of FA and MD without additional diffusion scanning. The synthesized diffusion features preserve biologically meaningful information and, when combined with T1-weighted MRI, improve AD diagnosis, particularly for MCI. By reducing scan time while retaining the benefits of multi-modality imaging, this approach offers a practical and scalable pathway to more accessible and accurate AD diagnosis in clinical settings.

\section*{Acknowledgements}
I used de-identified data from the Human Connectome Project(HCP), WU-Minn Consortium (Prin-cipal Investigators: David Van Essen and Kamil Ugurbil; NIH grant 1U54MH091657), funded by the 16 NIH Institutes and Centers that support the NIH Blueprint for Neuroscience Research. Access requires agreement to the HCP Open Access Data Use Terms. T1-weighted MRI and diffusion-derived FA/MD maps from the HCP S1200 release were used in this study. All analyses comply with HCP policies. For full dataset access and terms of use, visit https://www.humanconnectome.org/.

\bibliographystyle{unsrt}
\bibliography{references}

\appendix
\renewcommand{\thesection}{Appendix \Alph{section}}

\section{Ablation Studies}

\begin{table}[!ht]
\caption{Ablation Metrics for FA and MD Predictions.}
\vspace{8pt} 
\centering
\begin{tabular}{|c|c|c|c|c|c|c|}
\hline
\multicolumn{2}{|c|}{\textbf{Ablation}} & \textbf{MAE} & \textbf{MSE} & \textbf{RMSE} & \textbf{SSIM} & \textbf{Pearson}  \\ \hline
\multirow{2}{*}{\textbf{Proposed model }} & FA & 0.06652 & 0.01348 & 0.11454 & 0.64759 & 0.82687 \\ 
                                     & MD & 0.04747 & 0.00841 & 0.09093 & 0.71258 & 0.88179 \\  \hline
Single-task & FA & 0.06666 & 0.01357 & 0.11488 & 0.63831 & 0.82684 \\   \hline
Single-task & MD & 0.04676 & 0.00785 & 0.08814 & 0.70397 & 0.89028 \\  \hline
\multirow{2}{*}{Base Channel (16)} & FA  & 0.07569 & 0.01790 & 0.12837 & 0.78522 & 0.55883 \\ 
                                    & MD & 0.07274 & 0.05554 & 0.16152 & 0.71559 & 0.44838 \\  \hline
\multirow{2}{*}{Base Channel (64)} & FA  & 0.08308 & 0.01982 & 0.13565 & 0.75717 & 0.50422 \\ 
                                    & MD & 0.07646 & 0.02014 & 0.13633 & 0.73308 & 0.46780 \\  \hline
\multirow{2}{*}{Patch 32$^3$} & FA & 0.05380 & 0.01290 & 0.10416 & nan & 0.71392 \\  
                           & MD & 0.03876 & 0.00823 & 0.08409 & nan & 0.77777 \\  \hline
\multirow{2}{*}{Patch 96$^3$} & FA & 0.04855 & 0.00900 & 0.09453 & 0.88482 & 0.73565 \\  
                           & MD & 0.03339 & 0.00514 & 0.07154 & 0.93019 & 0.80629 \\  \hline
\end{tabular}
\label{tab:metrics21}
\end{table}

\subsection{Multi-task vs Single-task Prediction}
I tested whether predicting FA and MD jointly would help the model learn more 
useful representations. The multi-task model predicts FA and MD simultaneously 
which allows shared encoder features to capture common diffusion-relevant anatomical 
patterns and improves accuracy for both tasks\cite{26, 27}.

\subsection{Model Width}
Base channel numbers of 64, 32 (main model), and 16 were tested. Higher channels improve 
feature richness but consume more GPU memory; lower channels reduce computational 
cost but may underfit \cite{26}.

\subsection{Patch Size}
Patch sizes of 32$^3$, 64$^3$ (main method), 96$^3$, Larger patches improved anatomical context 
and benefit FA/MD prediction with higher memory consumption. The patch size of 64$^3$ can capture 
sufficient anatomical context to learn meaningful spatial patterns and maintain manageable 
GPU memory consumption. Smaller patches (e.g., 32$^3$) may lose global structural information 
and reduce prediction accuracy, whereas larger patches (e.g., 96$^3$) increase memory demands 
without significant performance gain.

\section{Model Diagrams}

\begin{figure}[!ht]
    \centering
    \includegraphics[width=0.8\textwidth]{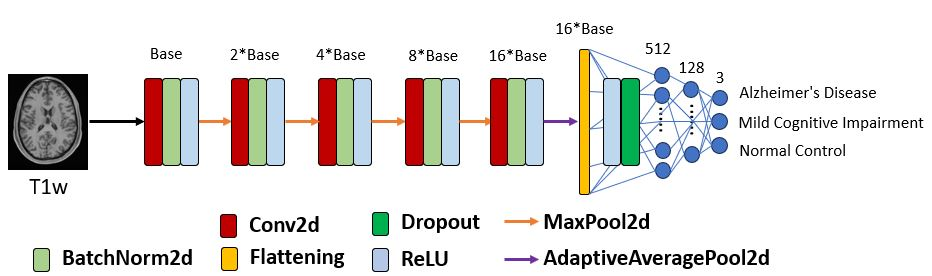}
    \caption{\textbf{Three Classification Single Modality Prediction Model}}
\end{figure}

\begin{figure}[!ht]
    \centering
    \includegraphics[width=0.8\textwidth]{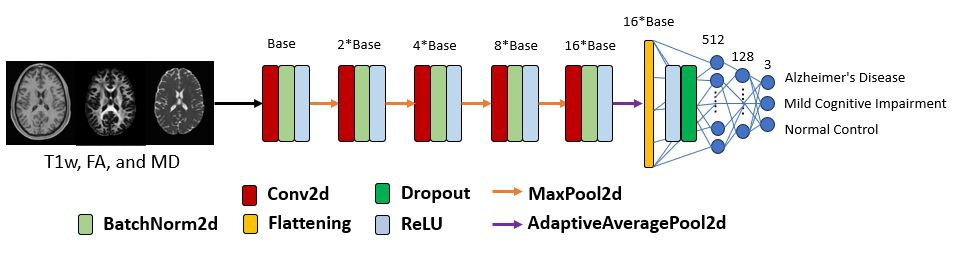}
    \caption{\textbf{Three Classification Multi-Modality Prediction Model}}
\end{figure}

\begin{figure}[H]
    \centering
    \includegraphics[width=0.8\textwidth]{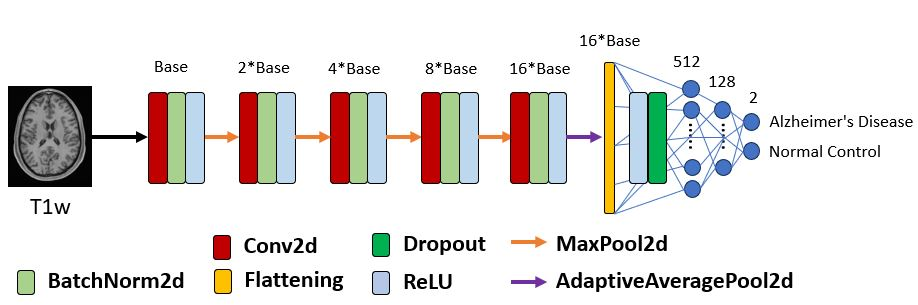}
    \caption{\textbf{Binary Classification Single Modality Prediction Model}}
\end{figure}

\begin{figure} [H]
    \centering
    \includegraphics[width=0.8\textwidth]{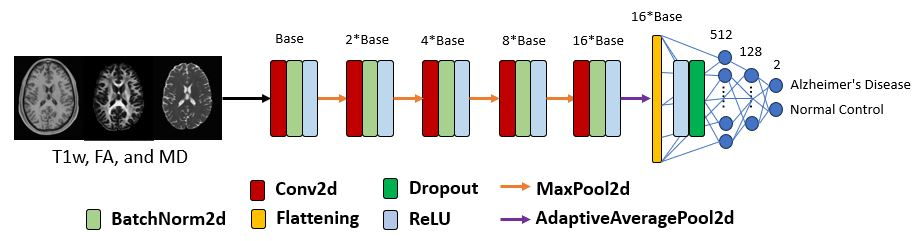}
    \caption{\textbf{Binary Classification Multi-Modality Prediction Model}}
\end{figure}

\end{document}